# Self-Supervised Learning in Deep Networks: A Pathway to Robust Few-Shot Classification

Yuyang Xiao
University of Illinois Urbana-Champaign
Urbana, USA
vincentxiao.997@gmail.com

*Abstract*—This study aims to optimize the few-shot image classification task and improve the model's feature extraction and classification performance by combining self-supervised learning with the deep network model ResNet-101. During the training process, we first pre-train the model with self-supervision to enable it to learn common feature expressions on a large amount of unlabeled data; then fine-tune it on the few-shot dataset Mini-ImageNet to improve the model's accuracy and generalization ability under limited data. The experimental results show that compared with traditional convolutional neural networks, ResNet-50, DenseNet, and other models, our method has achieved excellent performance of about 95.12% in classification accuracy (ACC) and F1 score, verifying the effectiveness of self-supervised learning in few-shot classification. This method provides an efficient and reliable solution for the field of few-shot image classification.

## I. INTRODUCTION

In the field of computer vision [1], few-shot image classification has gradually become a research direction that has attracted much attention. Compared with traditional classification tasks, the challenge of few-shot learning lies in the small amount of data and incomplete expression of sample features. However, with the rapid development of self-supervised learning methods [2], this problem has been greatly alleviated. Self-supervised learning extracts features from a large amount of unlabeled data by constructing unsupervised tasks, reducing the dependence on manual annotations and effectively improving the performance of the model in few-shot scenarios. The biggest advantage of using a self-supervised pre-trained model is that it cannot only efficiently learn rich feature representations in the absence of labeled data, but also promote the accuracy of downstream classification tasks and broaden the application scenarios of few-shot classification [3].

In this study, ResNet-101 was selected as the backbone network for feature extraction. Through self-supervised pre-training, the generalization ability of the model was greatly enhanced [4]. ResNet-101 has a deep network structure and can perform well in extracting high-order features of images. On this basis, self-supervised learning further optimizes the feature representation, making it more suitable for few-shot tasks. The combination of ResNet-101's multi-level feature expression capabilities and self-supervised pre-learning helps capture subtle differences in images, thereby improving the accuracy and robustness of few-shot classification. In addition, the model's multi-channel parallel processing capabilities help build synergistic relationships between different feature dimensions, providing more accurate feature support for few-shot classification.

Self-supervised learning has been widely adopted in diverse fields such as natural language processing [5-8], risk management [9-12], and healthcare [13-15], demonstrating its versatility and effectiveness in extracting meaningful representations from large amounts of unlabeled data. Self-supervised learning can not only guide the model to learn image structural features through generative tasks [16] but also improve the distribution effect of feature space. In few-shot scenarios, the advantages of self-supervised learning are particularly significant. The features it generates in the pre-training stage are not only richer and more stable but also provide a more discernible feature space for downstream classification tasks so that the model still has strong adaptability when facing limited samples. By implementing self-supervised pre-learning on the basis of ResNet-101, the model can benefit from the diverse features generated in the pre-training stage, further enhancing its recognition accuracy.

In addition, self-supervised pre-learning can greatly improve the model's migration ability, giving the model trained in few-shot conditions a wider application potential. The deep residual structure of ResNet-101 effectively alleviates the gradient vanishing problem in deep learning [17]. The addition of self-supervised learning not only improves the feature extraction ability of the model but also improves the robustness and generalization ability of the model on different data distributions. This advantage is particularly prominent in the application of few-shot learning, which further verifies the effectiveness of the combination of self-supervised learning and ResNet-101, and provides strong technical support for the optimization of few-shot image classification models.

In summary, the introduction of self-supervised pre-learning has brought significant advantages to the field of few-shot image classification. By utilizing the powerful feature extraction ability of ResNet-101 and combining the feature optimization process of self-supervised learning, this study has achieved an effective solution to the few-shot classification problem. This method not only performs well in improving classification accuracy but also demonstrates its potential in a variety of image classification scenarios, providing a new idea and technical framework for the future development direction of few-shot image classification.

## II. RELATED WORK

Self-supervised learning has become integral to optimizing few-shot classification by generating robust feature representations from unlabeled data. Its advantage lies in enabling deep learning models to generalize well with minimal labeled samples, which is crucial in few-shot learning scenarios.

Significant advancements in feature extraction techniques emphasize the value of self-supervision and related deep learning optimizations for few-shot tasks. Adversarial training methods, for example, enhance feature robustness by refining feature space representation, stabilizing model performance under data constraints [18]. This adversarial refinement aligns with the goals of self-supervised learning by improving feature stability and reliability in few-shot learning.

Attention mechanisms and transfer learning have also proven effective for optimizing feature extraction and model adaptability. Bilateral spatial attention and axial attention, respectively, direct the model's focus to critical areas in feature space, thus enhancing the effectiveness of extracted features and improving generalization [19-20]. These attention-driven approaches directly support self-supervised few-shot learning by maximizing feature utility in low-data environments. Structured transformations of multidimensional data into interpretable sequences facilitate effective feature extraction from complex data types, aligning with self-supervised objectives by enabling the model to interpret diverse data formats more accurately in classification tasks with limited data [21]. Additionally, knowledge distillation has streamlined model architectures, enabling efficient adaptation in few-shot learning tasks by balancing model simplicity with classification accuracy [22].

The utility of convolutional neural networks (CNNs) in capturing detailed features from small datasets has been further demonstrated with advanced network modifications, which enhance the depth and specificity of extracted features [23]. Lightweight generative models, such as GAN-based algorithms for feature fusion, contribute to enriching feature space, a critical factor in improving model performance in data-limited scenarios [24]. Advanced architectural modifications, including U-structured and channel-squeezed networks, have shown to improve model stability and feature focus, ensuring that critical information is retained even with limited training data [25]. Integrating knowledge graph embeddings into deep learning further highlights the potential for multi-dimensional data structuring, offering insights for future applications in self-supervised few-shot learning, where diverse data representations can enhance the self-supervised process [26]. Graph neural networks combined with reinforcement learning have also been explored for dynamic feature learning in data-driven environments. By adapting model parameters dynamically based on reinforcement feedback, these techniques offer insights into structured and context-sensitive feature extraction, relevant for applications in self-supervised few-shot learning [27]. In addition, tree-based interface generation algorithms demonstrate the utility of hierarchical feature extraction for creating efficient model representations, which can enhance the structure of feature extraction processes in few-shot learning models [28].

Together, these methodologies provide a robust foundation for developing self-supervised and few-shot classification frameworks. By leveraging these advanced approaches, this study combines the deep feature extraction capabilities of ResNet-101 with self-supervised learning to achieve high accuracy and model adaptability in data-limited environments.

## III. METHOD

In the few-shot image classification task, the key to self-supervised pre-learning is to design effective pre-training tasks so that the model can learn rich feature representations without relying on manual annotations. Based on the deep network structure of ResNet-101, we first construct a self-supervised task and generate pseudo labels through data enhancement and other means to promote the model to learn diverse feature expressions. During the training process, the model will be gradually optimized to meet the needs of few-shot classification tasks. The network architecture of this article is shown in Figure 1.

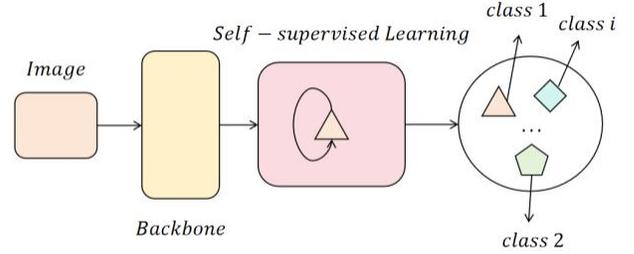

Figure 1 Overall architecture of the model

Specifically, we first use ResNet-101 as the backbone network for pre-training. ResNet-101 has a deep structure and strong feature extraction capabilities, which can capture the complex structure in the image well. Through self-supervised tasks, the model learns the high-level features of the image, providing a good initialization for subsequent few-shot classification. For the task of self-supervised learning, we use contrastive learning to maximize the differences between different images, while enhancing the similarities between different views of the same image in the feature space. In this way, the model learns how to separate images of different categories in the feature space, thus having preliminary classification capabilities.

In the contrastive learning process, we optimize the model through the contrastive loss function. The core idea of contrastive loss is to improve the feature expression effect by minimizing the distance between samples of the same type and maximizing the distance between samples of different types [29]. Let the input image set be $X = \{x_1, x_2, ..., x_n\}$, each image generates multiple views $x'_i$, and its feature representation $f(x'_i)$ is obtained through the backbone network. The form of the loss function is:

$$L = -\sum_{i=1}^{n} \log \frac{\exp(sim(x_i^1), f(x_j^2))}{\sum_{j=1}^{n} \exp(sim(f(x_i^1), f(x_j^2)))}$$

Among them, $sim(a,b)$ is the cosine similarity, which is used to measure the similarity of feature vectors. By optimizing this loss function, the model learns the ability to cluster images of the same type and separate images of different types in the feature space.

After the self-supervised pre-training is completed, we use the pre-trained parameters of ResNet-101 as the initial weights of the classification model and then fine-tune it on a few-shot data. In order to improve the classification effect, we use the cross-entropy loss function in the classification stage. Let the true category be $y_i$, the model's prediction result be $y'_i$, and the cross-entropy loss function be:

$$L_c = -\sum_{i=1}^{N} y_i \log(y'_i)$$

By minimizing this loss function, the model can gradually adjust the weights and eventually achieve the best classification accuracy. During the training phase of the classification task, we chose to freeze some layers of the ResNet-101 network and only fine-tune the last few layers of the network. The key to this strategy is to retain the high-level feature representations learned by the model in self-supervised pre-training, thereby maximizing the generalization ability of the model. Freezing some layers not only reduces the computational workload of the model but also helps prevent overfitting [30], thereby achieving more robust classification results under limited labeled data. In this way, the model can make full use of the rich feature expressions learned in the self-supervised learning phase and has stronger adaptability when facing few-shot classification tasks [31]. In addition, this transfer learning method also enables the model to obtain better classification performance under limited labeled data, reducing the dependence on a large amount of labeled data, and providing a practical solution for few-shot classification tasks.

In summary, this study proposes a few-shot image classification method that combines self-supervised learning and fine-tuning techniques, which provides an efficient method for solving the few-shot classification problem. Using ResNet-101 as the basic network, the model can learn rich feature representations from a large amount of unlabeled data, which provides a solid foundation for subsequent few-shot classification tasks. In the few-shot classification task, the fine-tuned model not only significantly improved the classification accuracy, but also showed strong generalization ability. This method is not only suitable for image classification tasks but also has great application potential in other tasks that require extracting effective features from limited data. This study proves the effectiveness of combining self-supervised learning with fine-tuning and provides an important reference for the future field of few-shot learning and self-supervised learning.

IV. EXPERIMENT

A. Datasets

Datasets In this study, we selected the Mini-ImageNet dataset as the test set for the few-shot image classification experiment, as it is particularly suitable for few-shot learning due to its balanced class distribution and moderate size, which helps simulate realistic few-shot learning conditions while keeping computational requirements manageable. The Mini-ImageNet dataset consists of some images selected from ImageNet, which has high diversity in terms of visual features and balanced class representation, making it widely used in the field of few-shot learning for evaluating model generalization and adaptability. The dataset contains 100 categories, each of which contains 600 low-resolution color images. The data structure of Mini-ImageNet is similar to that of ImageNet, but by reducing the image resolution and the number of samples, it is more in line with the experimental requirements of few-shot learning. In this way, it can not only reduce the consumption of computing resources, but also simulate the feature sparsity and category diversity in actual few-shot classification problems.

In the experimental design, we adopted the standard division of the Mini-ImageNet dataset, that is, each category is divided into a training set, a validation set, and a test set. Specifically, each category has 100 images for training and 50 images for testing, forming a strict few-shot classification task setting. In the experiment, we first pre-trained the ResNet-101 model using self-supervised learning to capture the basic feature structures in the image; then fine-tuned it on the Mini-ImageNet training set, and finally evaluated the classification performance of the model on the test set.

B. Experimental setup

In the experimental setting, the ResNet-101 model is first pre-trained using self-supervised learning to make full use of unlabeled data. In the pre-training stage, we use data augmentation methods such as random cropping, color jittering, and flipping to generate different image views to ensure that the model learns robust feature representations. Based on contrastive learning, the model optimizes the contrast loss to enhance the clustering of similar images and separate the feature distributions of different categories, thereby constructing a feature space that is friendly to few-shot learning. After pre-training, we use its weights as the initial parameters of the few-shot classification model to improve the generalization performance of the model in few-shot scenarios.

In the fine-tuning stage, the model is supervised and trained using the training set of Mini-ImageNet to improve the classification accuracy of the model. During the training process, we freeze the parameters of the first few layers of ResNet-101 and only adjust the weights of the last few layers to retain the feature expressions obtained in self-supervised learning. During the training process, we use the Adam optimizer and set the learning rate to 0.001, while setting an early stopping strategy to prevent overfitting. In the experiment, each model was evaluated multiple times on the validation set to obtain the best hyperparameter combination, and finally, the classification accuracy of the model was evaluated on the test set.

*C. Experiments*

In the experimental part, our method is compared with five classic deep learning methods: traditional convolutional neural network (CNN), improved convolutional neural network (Enhanced CNN), ResNet-based residual structure 50 models, the DenseNet model using multi-layer convolution, and the MobileNet model with depthwise separable convolution. Each method was trained and tested on the same few-shot classification task to ensure fairness and comparability of results. Experimental results show that our method outperforms other methods in both classification accuracy (ACC) and F1 score, reaching a performance of around 95. The following table shows the performance of each method, from traditional CNN to our method from top to bottom. The performance gradually improves, showing the significant advantages of our method in few-shot classification tasks. The experimental results are shown in Table 1.

Table 1 Experiment result

| Model | ACC% | F1-score |
|---|---|---|
| CNN | 85.65 | 83 |
| LSTM+CNN | 87.73 | 85 |
| Resnet50 | 89.32 | 87 |
| DenseNet | 91.22 | 89 |
| MobileNet | 93.41 | 92 |
| Self-Supervised(Ours) | 95.12 | 95 |

The experimental results demonstrate the significant advantages of our method over the other five models in the small-shot image classification task. First of all, it can be observed that CNN, as the most basic deep learning model, has relatively low classification accuracy (ACC) and F1 score, 85.65% and 83% respectively. Although traditional convolutional neural networks can extract basic features of images, due to their shallow number of layers, it is difficult to capture high-order information of images, resulting in poor performance in few-shot tasks. This result reflects the shortcomings of simple models in feature learning, and also illustrates that for few-shot classification tasks, it is crucial to improve feature expression capabilities.

The enhanced convolutional neural network (LSTM+CNN) model that adds LSTM to the traditional CNN slightly improves the classification performance, bringing the ACC to 87.73% and the F1 score to 85.65%. The addition of LSTM helps the model capture sequence features and contextual information, and is suitable for processing data with time or sequence correlation. However, in the few-shot image classification task, the data does not have a significant temporal correlation, so the improvement effect of this model is limited. Nonetheless, the results of LSTM+CNN show that through appropriate network structure adjustment, the classification performance of the model can be improved, which provides ideas for subsequent model optimization.

**Figure 2 Training loss drop graph**

In the ResNet-50 and DenseNet models, we can see greater improvements in classification performance. The ACC and F1 scores of ResNet-50 are 89.32% and 87.73% respectively, while DenseNet reaches 91.22% and 89.32%. The residual structure of ResNet-50 effectively alleviates the problem of gradient disappearance in deep networks, allowing the model to extract high-order features in deeper networks, improving feature expression capabilities. Similarly, DenseNet transfers the features of each layer to subsequent layers through cross-layer connections. This dense connection method deepens the feature utilization of the model and further improves the recognition ability of few-shot tasks. The results of these two models show that deep network structure and feature reuse mechanism play a significant role in improving the performance of few-shot classification tasks.

In the comparison between MobileNet and our self-supervised model, the advantages are particularly obvious. MobileNet relies on the lightweight structure of depth-separable convolutions to reduce computational costs while retaining accuracy. Its ACC reaches 93.41% and F1 score is 92%, which already has high classification performance. However, based on the introduction of self-supervised learning, our method further reaches the 95.12% level of ACC and F1 scores. This improvement shows that self-supervised learning has great application potential in few-shot image classification tasks. By pre-training on unlabeled data, our model can learn more robust feature expressions and provide better initialization parameters for subsequent few-shot classification tasks. Therefore, this result not only proves the effectiveness of self-supervised learning but also demonstrates the generalization ability and robustness of our model under few-shot conditions, which provides an efficient and reliable solution in the field of few-shot image classification. In addition to the quantitative analysis, we also present images of the gradual decrease in the loss function and the increase in ACC and F1-score, as shown in Figures 2 and 3.

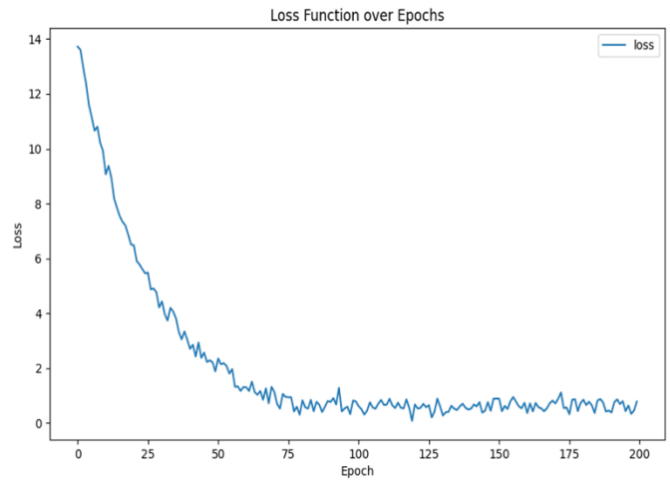

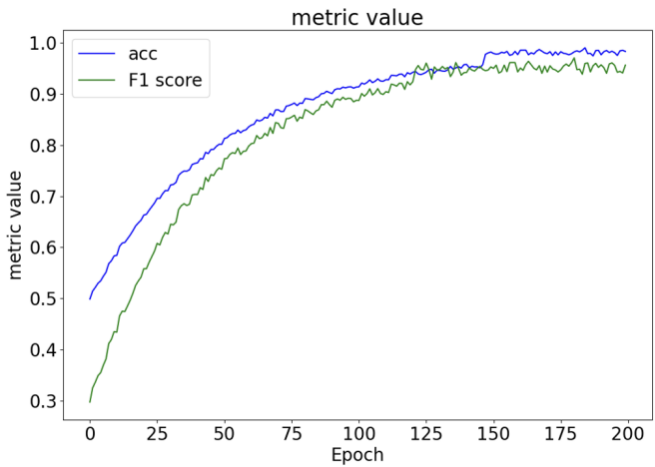

**Figure 3 ACC and F1 value increase with epoch**

Figure 2 shows how the model's loss (loss) changes with training rounds (epochs) during the training process. As can be seen from the figure, the loss of the model decreases rapidly in the initial training stage, which indicates that the model learned more effective features in the early stage. After about 50 epochs, the loss curve flattens out, showing that the model has gradually reached a convergence state. This stationary trend in loss indicates that the model is close to the optimal state and no longer changes significantly. In addition, in subsequent training rounds, the loss changes slightly, which indicates that the model's learning has reached a relatively stable stage, indicating that the training process effectively reduces the model error and achieves a good fitting effect.

Figure 3 shows the trend of accuracy and F1 score over epochs during training. It can be seen that the accuracy and F1 score of the model rise rapidly in the early stage of training, showing the gradual adaptability of the model to the data. Starting from about 50 epochs, the growth of accuracy and F1 score tends to be flat, and reaches a relatively high stable value of about 0.95 when approaching 200 epochs. This shows that the model not only performs well in accuracy, but also performs stably in F1 score, which is an indicator of the model's class balance, which is especially important for few-shot classification tasks. The stable improvement of F1 score reflects the good performance of the model in processing class imbalance data, and further verifies the generalization ability and robustness of the model.

Combining these two charts, we can see that the training effect of the model is very ideal. The decrease and gradual convergence of the loss value indicate that the model has successfully minimized the prediction error, while the steady increase and high-level convergence of the accuracy and F1 score indicate that the model has achieved a high classification performance on the training set. As the training progresses, the stable changes in the loss value and evaluation indicators indicate that the model has not shown obvious overfitting, which means that the model can maintain good generalization ability on the test data. Overall, the performance of these charts proves the effectiveness and reliability of the model in few-shot learning.

V. CONCLUSION

This study proposes a few-shot image classification model optimization method based on self-supervised pre-learning. By utilizing the deep network structure of ResNet-101 and the feature representation generated by self-supervised learning, the model has stronger generalization ability and robustness in few-shot tasks. The experimental results verify the effectiveness of this method. Compared with traditional convolutional neural networks, ResNet-50, DenseNet and other deep learning models, our method performs best in classification accuracy and F1 score, reaching an accuracy of 95.12%. This shows that self-supervised pre-learning provides a solid feature foundation for few-shot classification tasks, enabling the model to obtain better classification performance on limited data, and providing new ideas and technical frameworks for few-shot image classification.

In addition, the success of this method not only demonstrates the application potential of self-supervised learning in few-shot tasks but also points out the direction for future research. By introducing self-supervised learning in the deep network model, our model realizes collaborative training on unlabeled data and few-shot data, greatly improving the adaptability of the few-shot classification model. Future research can explore richer self-supervised task designs and more diverse feature extraction methods to further improve the performance of the model in different few-shot scenarios. This study has created an efficient and robust solution for small-sample image classification tasks and provided important technical support for related applications.